\documentclass{article} % For LaTeX2e
\usepackage{iclr2023_conference,times}

\usepackage{amsmath,amsfonts,bm}

\def\eqref#1{equation~\ref{#1}}
\def\1{\bm{1}}

\def\mA{{\bm{A}}}

\def\mM{{\bm{M}}}

\def\mX{{\bm{X}}}

\def\mZ{{\bm{Z}}}

\DeclareMathAlphabet{\mathsfit}{\encodingdefault}{\sfdefault}{m}{sl}
\SetMathAlphabet{\mathsfit}{bold}{\encodingdefault}{\sfdefault}{bx}{n}

\def\gG{{\mathcal{G}}}

\def\sH{{\mathbb{H}}}

\def\sR{{\mathbb{R}}}
\def\sS{{\mathbb{S}}}
\def\sT{{\mathbb{T}}}

\def\sV{{\mathbb{V}}}

\def\emX{{X}}

\usepackage{hyperref}
\usepackage{url}
\usepackage{booktabs}           % professional-quality tables
\usepackage{multirow}           % tabular cells spanning multiple rows
\usepackage{amsfonts}           % blackboard math symbols
\usepackage{graphicx}           % figures
\usepackage{duckuments}         % sample images
\usepackage{natbib}
\usepackage{wrapfig,lipsum,booktabs}

\title{Improving Subgraph Representation Learning via Multi-View Augmentation}
\author{Yili Shen \And {Xiao Liu} \And {Cheng-Wei Ju} \And Jiaxu Yan \And Jun Yi \And Zhou Lin \And Hui Guan}

\iclrfinalcopy % Uncomment for camera-ready version, but NOT for submission.
\begin{document}

\maketitle

\begin{abstract}
Subgraph representation learning based on Graph Neural Network (GNN) has exhibited broad applications in scientific advancements, %chemistry and biology, 
such as predictions of molecular structure--property relationships and collective cellular function. %collaborative gene functions.
%molecule property prediction and gene collaborative function prediction. 
In particular, %Moreover, 
graph augmentation techniques have shown promising results in improving graph-based and node-based classification tasks. 
% \zhou{Are we only talking about the classification tasks still? I added it tentatively to the title.} 
Still, they have rarely been explored in the existing GNN-based subgraph representation learning studies. %\zhou{It is not used what kind of tasks? I don't think our tasks are classification tasks.} %literature.
In this study, %paper, 
we develop a novel multi-view augmentation mechanism to improve subgraph representation learning models and thus the accuracy of downstream prediction tasks. 
%The 
Our augmentation technique creates multiple variants of subgraphs and embeds these variants into the original graph to achieve %both 
highly improved training efficiency, scalability, and accuracy. %\zhou{I am trying to make your tense to present tense consistently.}
%high training efficiency, scalability, and improved accuracy.  
Benchmark experiments
%Experiments 
on several real-world biological and physiological datasets %medical datasets %subgraph benchmarks 
demonstrate the superiority of our proposed multi-view augmentation techniques in subgraph representation learning. 
% \zhou{I think we will come back to revise the title and the abstract after we finish the draft. The current abstract looks identical to ICML Workshop. There are two styles we can try here. First, we can try to treat this work as a follow-up of the previous work, in which case I think we need to refer to that work frequently. Second, we can try to ``pretend'' the ICML paper does not exist and just focus on our new stuff. I don't know the underlying rule in the field.}
%The abstract paragraph should be indented 1/2~inch (3~picas) on both left and
%right-hand margins. Use 10~point type, with a vertical spacing of 11~points.
%The word \textsc{Abstract} must be centered, in small caps, and in point size 12. Two
%line spaces precede the abstract. The abstract must be limited to one
%paragraph.
\end{abstract}

\section{Introduction}
\label{Introduction}
%Full proceedings papers can have up to 9 pages with unlimited pages for references and appendix.

Subgraph representation learning using Graph Neural Networks (GNNs) can be broadly applied to various subgraph-related tasks in many fields of science and technology. 
As an outstanding example, the PPI (Protein--Protein Interaction) network~\citep{biosnapnets} uses nodes, edges, and subgraphs to represent single proteins, their interactions, and the set of interacting proteins, respectively. 
GNNs can be used to predict the biological processes (PPI-BP), cell component (PPI-CC), and molecular function (PPI-MF) by classifying the functionality of a subgraph (i.e., a group of proteins) in the PPI network. 
Another example is to apply GNN to fragment-based quantum chemical theory where each fragment in a crystal or aggregate is a subgraph and subgraph representation learning can predict the quantitative interactions between different fragments.  
Although applying GNNs to subgraph-related tasks~\citep{alsentzer2020subgraph,kim2022efficient,wang2021glass} starts to draw some attention, none of them have implemented graph augmentation techniques to improve task accuracy.

This work presents a novel multi-view approach to augment graphs for improving accuracy of subgraph classification tasks. 
Inspired by the effectiveness of graph contrastive learning~\cite{hassani2020contrastive,zhu2020deep,you2020graph}, our basic idea is to create multiple views of a subgraph by augmenting it, learn the embedding for each of the view, and then combine the representations for predicting the label of the subgraph. 
The rationale behind it is that the augmented subgraphs (i.e., the multiple views) essentially form an ensemble, which could provide more robust signal in determining the properties of the subgraph. 

\begin{figure}[ht]
\begin{center}
\fbox{\begin{minipage}{0.97\textwidth} %\rule[-.5cm]{0cm}{4cm}
\includegraphics[width=\textwidth]{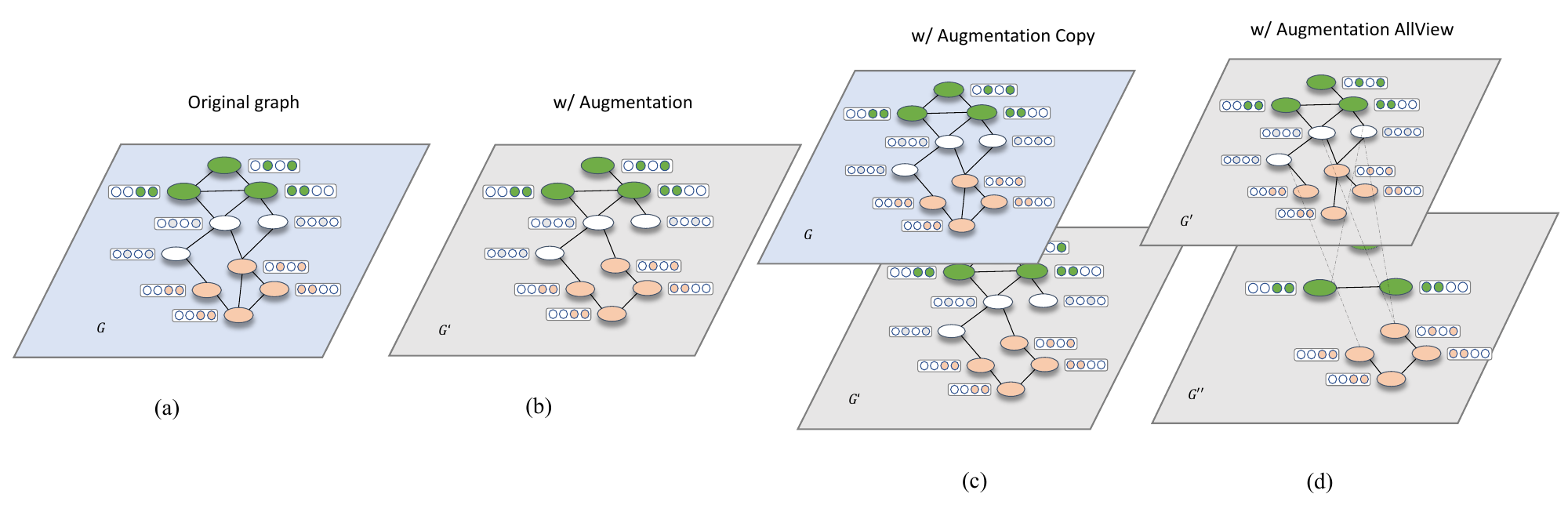}
\end{minipage}
%\rule[-.5cm]{4cm}{0cm}
}
\caption{Illustration of graph augmentation approaches. (a) The original graph $\gG$ contains two subgraphs (colored in orange and blue). (b) Augmented subgraphs are created by randomly dropping some edges in $\gG$. The new graph $\gG'$ is called the augmented graph. (c) A graph with two independent components where $\gG$ is the original graph and $\gG'$ is the augmented graph. Learning on this graph doubles the training cost.
(d) Augmented subgraphs $\gG"$ are embedded into an augmented graph $\gG'$.}
\label{fig:augmentation-approaches}
\end{center}
\end{figure}

The basic idea poses a fundamental challenge in how to efficiently create augmented subgraphs.  
Augmenting the entire graph to produce different views of the same subgraph is not scalable because the size of the augmented graph will grow linearly with the number of views.
Figure~\ref{fig:augmentation-approaches}(c) illustrates the problem. 
With only one additional view, GNNs need to conduct forward and backward propagations on two independent graphs (i.e., the original graph and the augmented graph)  during training, doubling the training cost. 
We address the efficiency issue by embedding augmented subgraphs in the original graph, significantly decreasing the demand for GPU resources. In this case, the computation of the embeddings for the augmented subgraphs can share intermediate representations within their neighborhood.
Figure~\ref{fig:augmentation-approaches}(d) illustrates an alternative efficient design where the augmented subgraphs are embedded into an augmented graph, instead of the original graph. 
We empirically validate that preserving the original view of subgraphs is essential for multi-view augmentation to improve task accuracy.

In summary, this work makes the following contributions:
\begin{itemize}
    \item This work proposes a novel multi-view augmentation strategy to improve the accuracy of subgraph-based learning tasks. This study is the first to
    explore the benefits of graph augmentation techniques in subgraph representation learning.
    \item The proposed multi-view augmentation strategy dynamically binds augmented subgraph views to the whole graph to drop exaggerated GPU resource consumption in order to achieve highly-improved training efficiency and task accuracy.  
    \item Empirical evaluations on three subgraph datasets demonstrate that our augmentation approach can improve existing subgraph representation learning by 0.3\%--2.9\% in accuracy, which is on average 1.1\% higher than general graph augmentation techniques DropEdge and GAug-M.
    % \hui{need to change the numbers}. 
 
\end{itemize}

\section{Related Works}

\paragraph{Subgraph Representation Learning}
Subgraph representation learning using GNNs has gained substantial attention these years~\citep{meng2018subgraph} due to its broad applications in scientific domains.  
Outstanding examples include SubGNN (SubGraph Neural Network)~\citep{alsentzer2020subgraph}, which routes messages for internal and border properties within sub-channels of each channel, including neighborhood, structure, and position. 
After that, %Then 
the anchor patch is sampled and the features of the anchor patch are aggregated to the connected components of the subgraph through six sub-channels. 
GLASS~\citep{wang2021glass} employs a labeling trick~\citep{zhang2021labeling} and labels nodes belonging to any subgraph to boost plain GNNs on subgraph tasks. S2N (Subgraph-To-Node)~\citep{kim2022efficient} translates subgraphs into nodes and thus reduces the scale of the input graph. These approaches focus on developing novel subgraph-based GNNs to improve task accuracy, but they have never implemented graph augmentation techniques.

\paragraph{Graph Augmentation} 
Data augmentation is a vital part of deep learning.
Many general graph augmentation techniques have been proposed to improve task accuracy recently. 
For node classification tasks, \cite{rong2020dropedge} proposes DropEdge to randomly drop the edges in a graph to enlarge the support of the training distribution.
DGI (Deep Graph Infomax)~\citep{velivckovic2018deep} perturbs the nodes by performing a row-wise swap of the input feature matrix while the adjacency matrix remains unchanged, generating negative samples for comparison learning and maximizing the mutual information of input and output.
GAug~\citep{zhao2021data} generates and removes edges of the graph by training an edge predictor to finally achieve the effect of high connectivity between nodes within %of 
the same class and low connectivity between nodes from %of 
different classes. 
NeuralSparse (Neural Sparsification)~\citep{zheng2020robust} proposes a supervised graph sparsification technique that improves generalization by learning to remove potentially task-irrelevant edges from the input graph.
GraphCL~\citep{you2020graph} points out that different data augmentation techniques introduce %bring 
different advantages in graph learning tasks in different domains. For example, edge perturbation can enhance %can be good for enhancing 
learning in social network graphs, but can be counterproductive in compound graphs learning %, edge perturbation can be counterproductive 
by destroying the original information. 
SUBG-CON (SUBGraph CONtrast)~\citep{jiao2020sub} samples a series of subgraphs containing regional neighbors from the original graph as training data to serve as an augmented node representation. 
Although these methods show promising results for augmenting graphs for node- and graph-based downstream tasks, they are not designed for augmenting subgraphs for subgraph-based tasks.

\paragraph{Multi-view Graph Learning}
Multi-view representation learning on graphs has attracted significant attention because they capture different properties on the same graph. 
Hassani {et al.}~\citep{hassani2020contrastive} introduce a multi-view graph learning manner to perform contrastive learning. 
O2MAC (One2Multi graph AutoenCoder)~\citep{o2mac} proposes a multi-view-based auto-encoder to promote self-supervised learning. 
MV-GNN (Multi-View Graph Neural Network)~\citep{DBLP:journals/corr/abs-2005-13607} utilizes two MPNNs (Message Passing Neural Networks)~\citep{DBLP:journals/corr/GilmerSRVD17} to encode atom and bond information respectively via multi-view graph construction. They construct multi-view graphs to express different levels of information in a graph, which is an intuitive and efficient way of building augmented graphs. 
Our work also leverages multi-view--based augmentation but focuses on subgraph-based tasks. 

\section{Notations and Preliminaries} %{The proposed method}

% \yili{Add some insights here.}

% \yili{Add the specific multi-view graph augmentation mathematical expression here.}
% \subsection{Preliminaries}
% \yili{I followed the dropedge preliminary section here, using some bold fonts to express different subsections of preliminaries.}
\subsection{Notations} 

Let $\gG = (\sV,\mathbb{E}, \mX)$ denote a graph, where $\sV=\{1, 2, .., N\}$ represents the node set, % and 
$\mathbb{E} \subseteq \sV \times \sV$ represents the edge sets, 
and $\mX$ {
% \color{red}
{is the matrix that}} represents the corresponding node feature. 
$\emX_{i}$, the $i^\text{th}$ row of $\mX$, represents the features associated with the $i^\text{th}$ node. Let $v_{i}$ denote a node in $\gG$.
The adjacency matrix $\mA\in \{0, 1\} ^{N\times N}$, where $a_{ij} = 1$ denotes that $(v_{i}, v_{j}) \in \mathbb{E}$.
% \zhou{What are your v's here? What happens to off-diagonal elements of $\mA$?}
% 
$\gG_S = (\sV_{S},\mathbb{E}_{S}, \mX_{S})$ denotes a subgraph of $\gG$, where $\sV_{S} \subseteq \sV$, $\mathbb{E}_{S} \subseteq \mathbb{E} \cap{(\sV_{S} \times \sV)_{S}}$, and $\mX_{S}$ stacks the rows of $\mX$ belonging to $\sV_{S}$.
The adjacency matrix of a subgraph $\gG_S$ is $\mA_{S}$.

\subsection{Subgraph Representation Learning}

% \yili{Splited the subgraph representation learning section}
Given the {
% \color{red}
{set of}} subgraphs $\sS = \{\gG_{S_{1}}, \gG_{S_{2}}, .., \gG_{S_{n}}\}$ and their labels $\sT=\{t_{S_{1}}, t_{S_{2}}, ..., t_{S_{n}}\}$, the goal of {\it Subgraph Representation Learning} is to learn a representation embedding $h_{S_{i}}$ for each subgraph $\gG_{S_{i}}$ to predict the corresponding $t_{S_{i}}$. 
% \zhou{$h$'s and $t$'s are numbers or functions or vectors?}

% \yili{Give GNN def here, follow the dropedge, brief}

% \yili{Introduce the motivation to use multi-view subgraph aug here.}
% \subsection{Motivation} 
% \yili{Show some privilege of whole-graph aug here: easy to implement. Show drawback: super large graph occupies too much gpu memory when the sparsity decreases}
% \subsubsection{Whole-graph Augmentation}

% \yili{Same as above, show some privilege of subgraph aug: only focusing on subgraph inner structure, which is the most effective ones. Drawback: this kind of unbalanced augmentation will damage the expressive power of gnns}

% \subsubsection{Subgraph Augmentation}
% \yili{I removed DropNode here since in the experiments we don't even used this. Also, I moved it from the section 'Proposed multiview augmentation' to section subgraph augmentation, since the logic flow goes better if I mention subgraph augmentation techniques in advance.}

\subsection{Graph Augmentation}

{
% \color{red}
{In the present work, we illustrate our multi-view augmentation scheme based on two typical existing graph augmentation strategies, DropEdge~\citep{rong2020dropedge} and GAug-M~\citep{zhao2021data}.}} 
%We also use two existing approaches, DropEdge~\citep{rong2020dropedge} and GAug-M~\citep{zhao2021data}, to illustrate typical graph augmentation strategies.  

DropEdge is a graph data perturbation strategy that randomly drops edges in a graph~\citep{rong2020dropedge} 
so that it enlarges the training support to improve the performance of GNNs on node-level tasks. 
% \zhou{I think you need to mention why the method works the way it works.}
% \hui{add citation}
We employ DropEdge for each subgraph to generate an augmented subgraph, by generating 
%For each subgraph, we generate 
a stochastic boolean mask $\mM_{p}\in \sR^{m\times m}$,  where $m$ is the number of nodes in the subgraph and $p$ represents the rate of dropping edges. 
The new adjacency matrix becomes
${\mA}_{S}' =  \mA_{S} -  \mM_{p} \odot \mA_{S}$,
where $\odot$ means the element-wise product. 
% \zhou{Is $\sR$ the set of real numbers?}

%\zhou{Is there a title for this paragraph?} 
GAug-M~\citep{zhao2021data} is a graph data augmentation strategy that leverages neural edge predictors to promote intra-class and demote inter-class edges so as to %that 
form new edge weights. It contains a two-stage training schema.
In the first step, %first using 
we use VGAE (Variational Graph Auto-Encoders)~\citep{kipf2016variational} as the edge predictor to get an edge-probability matrix $\mM$, which describes the graph's probabilistic connectivity, $\mM=\sigma(\mZ\mZ^T)$, where $\mZ = \text{GNN}(\mA, \mX))$.
% \zhou{Is $\mZ$ a matrix too?}
Denote $|\mathbb{E}|$ as the number of edges in graph $\gG$. 
In the second step, %Secondly, 
we use the probability matrix $\mM$, to make the top $i|\mathbb{E}|$ non-edges with the highest edge probabilities to be connected, and the least $j|\mathbb{E}|$ edges with the lowest edge probabilities to be disconnected to produce an augmented graph from $\gG$ to $\gG'$, where $i, j \in \{0, 1\}$.
% \zhou{The sentence above is not very readable...}

% \subsection{Graph Contrastive Learning}
% Graph contrastive learning (GraphCL)~\citep{you2020graph} utilized contrastive loss for self-supervised learning to do graph self-supervised learning. The contrastive loss is to enforce maximizing the consistency between positive pairs ${z_i}, {z_j}$  compared with negative pairs
% \yili{Refered graph contrastive loss} so that the agreement between two augmented views of the same graph can be maximized.

\section{Methods} %{Our Method}
\label{sec:method}

In this section, we present
our proposed multi-view augmentation approach, following by 
analyses of the computational complexity
and discussions on the shortage of the alternatives shown in Figure \ref{fig:augmentation-approaches} and the advance of our multi-view approach.
% \hui{fill in}. 

\subsection{Methodology} 

\begin{figure}[ht]
\begin{center}
\fbox{\begin{minipage}{0.97\textwidth} %\rule[-.5cm]{0cm}{4cm}
\includegraphics[width=\textwidth]{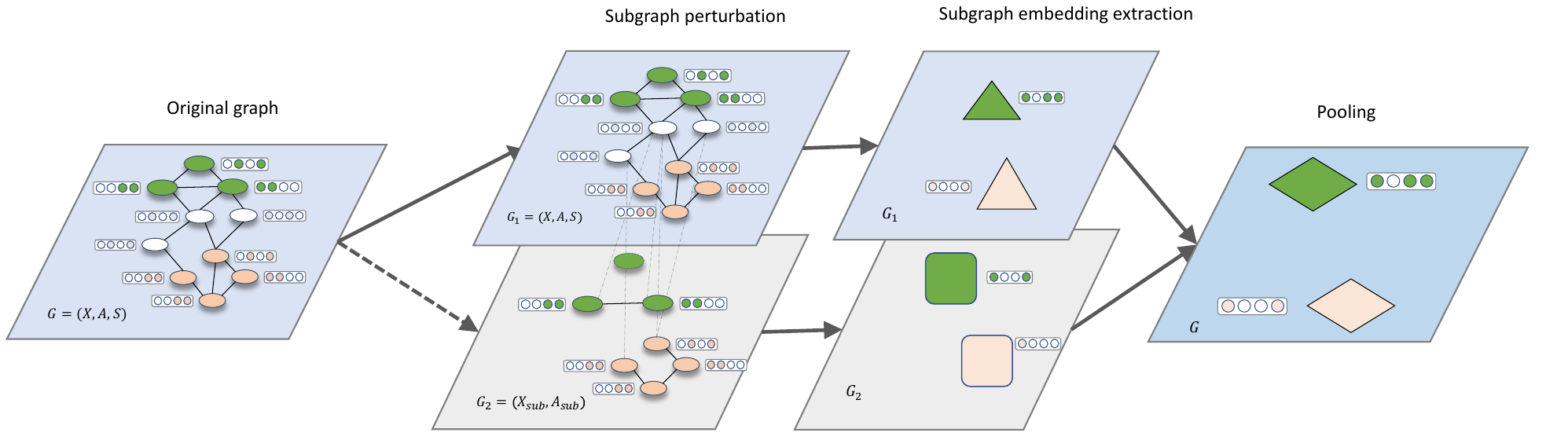}
\end{minipage}
%\rule[-.5cm]{4cm}{0cm}
}
\end{center}
\caption{Overview of our proposed subgraph augmentation approach. The two subgraphs in the original graph are colored in green and orange.  We first generate multi-subgraph views via stochastic augmentation. Following that we connect the augmented subgraph to the remaining part of the original graph, by adding edges that link the augmented subgraph and the whole graph. After feeding forward the whole graph into subgraph-specific GNNs, we extract the subgraph embeddings of different views, respectively (triangles and squares).  Ultimately, we fuse the embeddings of different views %are fused 
by a pooling function and obtain the augmented subgraph embeddings (diamonds). 
% \zhou{can we cut short the caption a little? It is even longer than the figure and has some repeating information with the text.}
}
\label{fig:overview}
\end{figure}

Figure~\ref{fig:overview} illustrates the basic idea of the multi-view augmentation {{that is implemented in the present study}}. 
At each forward step, we generate augmented views of subgraphs in this batch by randomly perturbing original subgraphs with a particular graph data augmentation strategy.
After that, we add the augmented subgraphs to the original graph and feed the new graph into a subgraph-specific neural network. 
Here, we obtain subgraph embeddings of both the original subgraph and the augmented subgraph. 
These embeddings are fed into a pooling function to generate a single subgraph embedding for each subgraph, which is used for downstream subgraph-based tasks. 
Meanwhile, to maximize the agreement between the original subgraph embeddings and the augmented subgraph embeddings, we utilize the contrastive loss between the original graph and augmented graphs.

After subgraph augmentation, we obtain 
% \yili{Replacing an augmented subgraph to augmented subgraphs since we have a bunch of them}%an augmented subgraph 
augmented subgraphs $\gG_S' = (\sV_S', \mathbb{E}_S', \mX_{S}')$. We enrich the original graph to include both the augmented subgraph and the original subgraph. The  enriched graph is thus called a {\it Multi-View Graph}.
Mathematically, the multi-view graph $\gG' = (\sV', \mathbb{E}', \mX')$ where $\sV' = \sV \cup \sV_S'$. 
The consequent adjacency matrix becomes 
\begin{equation}
    \mA'=
    \begin{bmatrix}
        \mA & \mA[:, \sV_S'] \\
        \mA[\sV_S', :] & \mA_S
    \end{bmatrix}.
\end{equation}

Feeding forward the multi-view graph into subgraph-specific neural networks, by selecting the subgraph and augmented subgraph nodes by masks ${M}_{S_O}, {M}_{S_A}$, we can get the embeddings of both the augmented subgraph and the original subgraph and denote them as $h_{S_{O}}$ and $h_{S_A}$, respectively. We fuse different subgraph embeddings into one embedding by applying a pooling function (e.g., MaxPool or AvgPool): 
% \zhou{Which pooling function are you using exactly? No reference?}
\begin{align}
    \sH & = \text{GNN}(\gG') \\
    h_{S_O}, h_{S_A} &= \sH[{\mM}_{S_O}], \sH[{\mM}_{S_A}], \\
    h_S &= \text{Pool}(h_{S_{O}}, h_{S_A})\label{eq:pool}
\end{align}
With the learned subgraph embeddings, we can predict the subgraph properties by applying a MLP (Multi-Layer Perception),~\citep{hastie2009elements}
% \yili{Added hat}
\begin{equation}
    \hat{t}_{S} = \text{softmax}(\text{MLP}\ (h_{S})).
\end{equation}

Meanwhile, we 
%consider
implement
the contrastive loss between the augmented views, which enforces the embeddings of the original subgraph and the generated subgraph embedding to be close and those of augmented subgraphs not generated by this subgraph to be distant. Let $\lambda$ be the coefficient to control the contrastive loss contribution, the total loss function becomes 
\begin{equation}
    \mathcal{L} = \mathcal{L}_\text{cls}({t_{S}}, \hat{t_{S}}) + \lambda\mathcal{L}_\text{contrast}(h_{S_{O}},  h_{S_A}).
\end{equation}
% \zhou{What do you mean by ``consider'' here? This verb is too weak. Do you mean ``implement''? Are ``close'' and ``distant'' the correct adjectives?}\yili{Here close and distance means the distance betweeen the embeddings. I immitated this part from GraphCL}

\subsection{Computational Complexity Analysis} 

The proposed subgraph multi-view augmentation is independent to %is unrelated to
augmentation strategies or subgraph-specific GNNs. Thus, we can train the model in an end-to-end fashion, which means there's neither modification at the beginning of each epoch nor after the forward. We can simply analyze %consider 
the additional computational complexity made by the augmentation, the corresponding augmented graph inference at each forward step.

Let the number of nodes in the original graph be $|\sV|$, the number of edges be $|\mathbb{E}|$, the number of the $i^\text{th}$ subgraph node be $|\sV_{S_i}|$, the number of the $i^\text{th}$ subgraph node be $|\mathbb{E}_{S_i}|$, and the training setting for batch size be $b$. Therefore, the expectation of $|\sV_{S}|$, $\mathop{\mathbb{E}}[|\sV_{S}|]$ is going to be 
$\mathop{\mathbb{E}}[|\sV_{S}|] = \dfrac{\sum_{i}|\sV_{S_i}|}{b}$. Likewise, the expectation of $|\mathbb{E}_{S}|$, $\mathop{\mathbb{E}}[|\mathbb{E}_{S}|]$ is going to be 
$\mathop{\mathbb{E}}[|\mathbb{E}_{S}|] = \dfrac{\sum_{i}|\mathbb{E}_{S_i}|}{b}$. 
% \zhou{Can you use a different number for the batch size?}\yili{I modified it to be $b$}

Considering a graph neural network for which the computational complexity is $O(f_\text{GNN}(|\sV|, |\mathbb{E}|))$, and the memory complexity is $O(g_\text{GNN}(|\sV|, |\mathbb{E}|))$, the computational complexity for an augmentation strategy is %. And an augmentation strategy whose  computational complexity is 
$O(f_\text{Aug}(|\sV|, |\mathbb{E}|))$, and the memory complexity is $O(g_\text{Aug}(|\sV|, |\mathbb{E}|))$.
Because we augment on subgraphs only during %During 
the augmentation stage, %since we are going to augment on subgraphs only, 
the computational complexity is %will be 
$O(f_\text{Aug}( \mathop{\mathbb{E}}[|\sV_{S}|], |\mathop{\mathbb{E}}[|\mathbb{E}_{S}|]))$, and %. Also, 
the memory complexity is %will be 
$O( g_\text{Aug}(\mathop{\mathbb{E}}[|\sV_{S}|], \mathop{\mathbb{E}}[|\mathbb{E}_{S}|]))$
With regard to the forward step, the total {
% \color{red}
{computational complexity}} of the whole graph is %will be 
$|\sV| + \mathop{\mathbb{E}}[|\sV_{S}|]$ and %the total number 
that of edges is % will be 
$|\mathbb{E}| + \mathop{\mathbb{E}}[|\mathbb{E}_{S}|]$. In this way, the inference computational complexity is %will be 
$O(f_\text{GNN}(|\mathbb{E}| + \mathop{\mathbb{E}}[|\mathbb{E}_{S}|], |\mathbb{E}| + \mathop{\mathbb{E}}[|\mathbb{E}_{S}|]))$, and the memory complexity is %will be 
$O(g_\text{GNN}(|\mathbb{E}| + \mathop{\mathbb{E}}[|\mathbb{E}_{S}|], |\mathbb{E}| + \mathop{\mathbb{E}}[|\mathbb{E}_{S}|]))$.

%Thus, 
To summarize our analysis, the overall computational complexity is 
$
    O( f_\text{Aug}( \mathop{\mathbb{E}}[|\sV_{S}|], |\mathop{\mathbb{E}}[|\mathbb{E}_{S}|])+ f_\text{GNN}(|\mathbb{E}| + \mathop{\mathbb{E}}[|\mathbb{E}_{S}|], |\mathbb{E}| + \mathop{\mathbb{E}}[|\mathbb{E}_{S}|])),
$
and the overall memory complexity is 
$
    O( g_\text{Aug}( \mathop{\mathbb{E}}[|\sV_{S}|], |\mathop{\mathbb{E}}[|\mathbb{E}_{S}|])+ g_\text{GNN}(|\mathbb{E}| + \mathop{\mathbb{E}}[|\mathbb{E}_{S}|], |\mathbb{E}| + \mathop{\mathbb{E}}[|\mathbb{E}_{S}|])).
$

% At the augmentation stage, 

% Considering a plain GNN as the subgraph-specific neural network, whose computational complexity is $O(T(|V|))$, and memory complexity is $O(S(|V|))$. 

% During training, first, the subgraphs go through an augmentation layer. Let the augmentation time complexity to be $O(f(|\mathcal{V}|))$. In the full-batch augmentation setting, the time complexity of augmentation is supposed to be $O(\sum_{s'}f(|\mathcal{V}_{s'}|))$. Meanwhile, subgraph multi-view augmentation has an extra space complexity of $O(\sum_{s'}|\mathcal{V}_{s'}|^2 + 2(\sum_{s'}|\mathcal{V}_{s'}|)|\mathcal{V}|)$ to save the extra edges, since we will create one view for each subgraph. In the GNN inference part, as the nodes are appended from $|\mathcal{V}|$ to $(\sum_{s'}|\mathcal{V}_{s'}| + |\mathcal{V}|)$, and GNN take an $O(|\mathcal{V}|)^2)$ time complexity, the corresponding time complexity becomes $O((\sum_{s'}|\mathcal{V}_{s'}| + |\mathcal{V}|)^2)$. Thus, the overall time complexity will be $O(\sum_{s'}f(|\mathcal{V}_{s'}|) +(\sum_{s'}|\mathcal{V}_{s'}| + |\mathcal{V}|)^2)$ and the overall space complexity will be $O(\sum_{s'}|\mathcal{V}_{s'}|^2 + 2(\sum_{s'}|\mathcal{V}_{s'}|)|\mathcal{V}| + |\mathcal{V}|^2)$.
% \yili{I don't know whether this paragraph goes correct, please kindly check it. And I also want to supplement that sampling trick to relieve the problem.}
%But fortunately, we adopt the sampling method..

\subsection{Strengths of Our Multi-View augmentation Scheme}
\subsubsection{Multi-View on subgraph}
% \zhou{Are you discussing different methods or our method? I think the title is not clear enough. You need to provide citations to prove your idea, if you have no mathematical proofs.}

\textbf{Augmentation on whole-graph vs. Augmentation on subgraph}\label{subgraphormulti} The augmentation strategies are supposed to directly modify the whole-graph edges and nodes. However, 
the time complexity becomes unaffordable for the augmentation strategy when it is greater than $O(|\sV|)$.
%when the augmentation strategy's time complexity is larger than $O(|\sV|)$, the time complexity is unaffordable. 
%Also, since 
Because the nodes of the subgraphs %nodes only 
only accept messages from their $k$-hop neighbors, it's more  efficient to augment subgraph-related edges.
% \yili{I want to say subgraph augmentation is as powerful as a whole graph setting. But I don't know how to prove that.}
% \zhou{I think you probably need to explain why it is more effective and efficient. If you cannot prove it in mathematics, you can provide a citation maybe?}

\textbf{Single-view vs. Multi-view} DropEdge~\citep{rong2020dropedge} and GAug-M~\citep{{zhao2021data}} take the strategy of single-view augmentation. One way to augment the subgraph-based data is to augment the original graph using the single-view augmentation on subgraph edges to perturb the message-passing flow of the subgraph, which will reduce the {
% \color{red}
{amount of the}} information contained in the original graph. Our {
% \color{red}
{multi-view}} method, on the other hand, can keep the original subgraph structures because it just modifies the augmented subgraph. 

\textbf{Copy graph vs. Subgraph multi-view} GraphCL~\citep{you2020graph} creates another graph view 
% \zhou{What do you mean by ``view'' here?} 
to allow %make 
contrastive samples to perform %do 
the self-supervised training. It's intuitive to create another whole graph by using the augmentation strategies. Neverthless, %such kind of 
such an augmentation flow doubles the %will require double 
memory and time consumptions, which is not affordable for large-scale graph learning tasks like social networks or protein-protein interactions. In this way, multi-view graph learning improves the hardware resource requirements and accelerates the computations.
% \zhou{Citations? citations?}

\textbf{Augmentation on all views vs. Preserving one original view} Although augmentation on all views will provide wider training support than keeping the original subgraph structures, however, there is a loss of the information from the original subgraphs. Because we perform the augmentation on the newly-created view, it's enough for perturbing so that keeping one original view can help the model to learn lossless graph-based information. 
In this way, our proposed multi-view augmentation can show more generalizability and efficiency in subgraph representation learning.
%As subgraph 
% \subsubsection{Multi-view augmentation on subgraph vs. Single-view augmentation on subgraph}

% % \zhou{Can we say Multi-view augmentation on subgraph vs. Single-view augmentation on subgraph?}

% Each subgraph can be treated as a graph, so it's intuitive to apply the augmentation strategy to each subgraph. However, the unbalanced augmentation will damage the expressive power of GNNs since it perturbs the consistency between subgraph-relevant edges and non-subgraph-relevant edges. Our approach at least keeps one view that can convey enough information from the unaugmented subgraph.

\subsubsection{Straight Inference vs. Joining in Contrastive Loss}
Comparing with straight inference, a contrastive loss can effectively make the source subgraphs and the augmented subgraphs closer and the other augmented subgraphs more distant. This strategy will prevent the increased disagreement of views after epochs of training.
%\chengwei{a little short, maybe some more sentence.}
% \yili{same, need a prove, but idk how to prove. }
% \zhou{Provide citations maybe?}
% \subsubsection{our approach vs other aug}
% \subsubsection{Augmentation on all views vs Keeping at least one view}

\section{Experiments}

In this section,  we evaluate the performance of our proposed subgraph multi-view augmentation strategy across a variety of architectures, datasets, and graph augmentation strategies, %and comparing with 
and compare them with other strategies to show the effectiveness of our multi-view subgraph augmentation.

\subsection{Experiment Settings}

\subsubsection{Datasets} 

\begin{wraptable}[6]{r}{0.62\textwidth}
\begin{minipage}{0.60\textwidth}
\vspace{-0.8cm}
\caption{Statistics of four real-world datasets.}
%\tabcolsep=0.06cm 
%\small 
\label{tab:datasets}
\begin{center}
\vspace{-0.4cm}
\begin{tabular}{lcccc}
%\toprule
%\hline
\multicolumn{1}{c}{\bf DATASET}  &\multicolumn{1}{c}{\bf NODES} &\multicolumn{1}{c}{\bf EDGES} &\multicolumn{1}{c}{\bf SUBGRAPHS}\\
%Dataset  & \#Nodes & \#Edges & \#Subgraphs
\hline %\midrule 
PPI-BP    & 17,080& 316,951& 1,591\\
HPO-METAB    &14,587& 3,238,174& 2,400\\
HPO-NEURO    &14,587& 3,238,174& 4,000\\
% EM-USER    &57,333&4,573,417& 324\\
\hline%\bottomrule 
\end{tabular}
\end{center}
\end{minipage}
\end{wraptable}

Table~\ref{tab:datasets} summarizes statistics of the datasets obtained from SubGNN~\citep{alsentzer2020subgraph}.
We follow the split reported in~\citet{wang2021glass}. 
Specifically,
PPI-BP~\citep{biosnapnets} aims to predict the collective cellular function of a given set of genes known to be associated with specific BP (Biological Processes) in common. 
The graph shows the correlation of the human PPI (Protein--Protein Interaction) %
network where nodes represent proteins and edges represent the interaction between proteins. 
A subgraph is defined by the collaboration of proteins and labeled according to cellular functions from six categories (metabolism, development, signal transduction, stress/death, cell organization, and transport). 
HPO-METAB and HPO-NEURO~\citep{splinter2018effect, hartley2020new} simulate rare disease diagnosis with the task of predicting subcategories of metabolic and {
% \color{red}
{neurological}} disorders that are the most consistent with these phenotypes. 
The graph is a knowledge graph containing phenotypic and genotypic information for rare diseases. 
A subgraph consists of a collection of phenotypes associated with rare monogenic diseases. 

\subsubsection{GNN Models}

The proposed augmentation technique is model-agnostic because it does not alter the GNN model. 
In the experimental part, we evaluate the proposed multi-view augmentation using a subgraph-specific model GLASS~\citep{wang2021glass} and two widely-used GNN architectures: GCN (Graph Convolutional Networks)~\citep{kipf2016semi} and GSAGE (GraphSAGE)~\citep{hamilton2017inductive}. % They are referred as GLASS, GCN, GSAGE, respectively.
We chose DropEdge~\citep{rong2020dropedge} and GAug-M~\citep{zhao2021data} to create subgraph views that are different from the orignal subgraph. 
% \zhou{What do you mean by evaluated and evaluation here?}

\subsubsection{Baselines for Comparison}
We compare our proposed approach {\bf  w/ Augmentation MV} with the following baselines.  
\begin{itemize}
    \item {\bf Original}: It directly trains the GNN model (i.e., GLASS, GCN, and GSAGE) to establish a baseline without any augmentation (see Figure~\ref{fig:augmentation-approaches}(a)). 
    \item {\bf  w/ Augmentation}: It applies DropEdge or GAug-M on the original graph $\gG$ to get an augmented graph $\gG'$ (see Figure~\ref{fig:augmentation-approaches}(b)). The GNNs will learn on the augmented graph  $\gG'$.  
    \item {\bf  w/ Augmentation Copy}: It applies DropEdge or GAug-M on the original graph $\gG$ to get an augmented graph $\gG'$ (see Figure~\ref{fig:augmentation-approaches}(c)). The GNNs will learn on both $\gG$ and $\gG'$. The final embedding is generated after a pooling operation as in Equation~\ref{eq:pool}. 
    \item {\bf  w/ Augmentation AllView}: It applies DropEdge or GAug-M on both original subgraphs and the augmented subgraphs, to generate $\gG'$ and $\gG''$, which fails to preserve the subgraph structure (see Figure~\ref{fig:augmentation-approaches}(d)). The final embedding is generated after a pooling operation as in Equation~\ref{eq:pool}. 
\end{itemize}

\subsubsection{Implementation Details}

We perform a hyperparameter search for all of the baselines and our approach and report the best mean F1 score on all test datasets. 
The searching space of hyperparameters and details are provided in 
% \yili{Add table here}
in the Appendix~\ref{sec:appendix}. 
% \zhou{Don't forget it!}
Because GAug-M~\citep{zhao2021data} requires an initial edge probabilistic distribution, we use VGAE~\citep{kipf2016variational} to perform %do 
the self-supervised learning on edge prediction to generate the distribution.
For the GNN model GLASS~\citep{wang2021glass}, we first train the model in an unsupervised manner, and then use supervision from downstream tasks to fine-tune the model hyperparameters, following the procedure provided by the original paper. %original paper's setting.
We perform 10 independent %different 
training and validation processes with 10 distinct random seeds.

\subsection{Results}

\subsubsection{Overall Results}

The empirical performance is summarized in Tables~\ref{tab:results_dropedge} and \ref{tab:results_gaugm}. 
Our proposed subgraph augmentation improves most of the task accuracy and improves across all three datasets where the whole-graph augmentation strategy 
% \yili{Modified here.}
succeeded in improving the accuracy, %from biological science and physiological science, indicating the universality of our strategy. 
we applied our strategy into GCN, GSAGE, and GLASS. It 
%consistently
performs better than most baseline approaches, mainly because it inhibits over-smoothing and over-fitting problems. 
% \zhou{I don't think Prof. Guan loves this sentence....}
Specifically, our approach improves the Micro-F1 scores by 0.6\%--2.9\%, 0.5\%--2.5\%, and 0.3\%--1.7\% compared to plain GLASS, GCN, and GSAGE, respectively.
%which are state-of-the-art approaches for subgraph representation learning. 
Specifically, when we apply multi-view augmentation (MV) to GCN and GSAGE, {
% \color{red}
{
we find that such backbones combined with multi-view augmentation show a great increase in performance.}}
% \zhou{I have not been able to make sense from this paragraph.}

\begin{table*}[ht]
\begin{center}
%\begin{small}
\caption{Mean Micro-F1 scores with standard deviations 
of the mean on three real-world datasets. %datasets. 
Results are provided from %with 
runs with 10 random seeds on DropEdge.}
\label{tab:results_dropedge}
\vspace{0.2cm}
\begin{tabular}{lcccc}
%\toprule
\multicolumn{1}{c}{\bf BACKBONE}  &\multicolumn{1}{c}{\bf METHOD} &\multicolumn{1}{c}{\bf PPI-BP}  &\multicolumn{1}{c}{\bf HPO-METAB}&\multicolumn{1}{c}{\bf HPO-NEURO}\\
\hline
%\midrule
\multirow{6}*{GLASS}  & {Original} & $0.610\pm0.006$  & $0.600\pm0.003$  & $0.678\pm0.004$  \\
\cline{2-5}
  & {w/ DropEdge} & $0.626\pm0.006$ & ${\bf 0.607}\pm0.008$ & $0.678\pm0.004$  \\
  % & {w/ DropEdge sub} & 0.583\pm0.006 & 0.584\pm0.009 & 0.651\pm0.005 \\
  & {w/ DropEdge copy} & $0.605\pm0.006$ & $0.593\pm0.013$ & $0.676\pm0.003$ \\
    & {w/ DropEdge AllView} & $0.613\pm0.007$ & $0.577\pm0.008$ & $0.672\pm0.006$ \\
   % & {w/ DropEdge whole} &  0.912\pm0.005 & {\bf 0.607 }\pm0.008 & 0.678\pm0.004 & {\bf 0.626}\pm0.006\\
   & {w/ DropEdge MV} & ${\bf 0.628}\pm0.005$ & ${\bf 0.607}\pm0.008$ & ${\bf 0.685}\pm0.003$ \\
  % \cline{2-5}
\hline %  \bottomrule
\multirow{6}*{GCN}  & {Original} &
$0.613\pm0.008$ & $0.553\pm0.018$ & $0.658\pm0.007$  \\
\cline{2-5}
& {w/ DropEdge} &
 $0.618\pm0.006$ & $0.556\pm0.006$ & $0.651\pm0.006$ \\
  % & {w/ DropEdge sub} & 0.583\pm0.006 & 0.584\pm0.009 & 0.651\pm0.005 \\
  & {w/ DropEdge copy} & $0.553\pm0.005$ & $0.349\pm0.026$ & $0.317\pm0.018$ \\
    & {w/ DropEdge AllView} & $0.595\pm0.006$ & $0.552\pm0.019$ & $0.613\pm0.010$ \\
   % & {w/ DropEdge whole} &  {\bf 0.884}\pm0.011 & {\bf 0.593}\pm0.007 & {\bf 0.692}\pm0.002 & 0.613\pm0.006\\
   & {w/ DropEdge MV} & ${\bf 0.619}\pm0.006$ & ${\bf 0.567}\pm0.006$ & $0.622\pm0.007$ \\
\hline % \bottomrule
  \multirow{6}*{GSAGE}  & {Original} &
 $0.621\pm0.006$ & $0.581\pm0.008$ & $0.684\pm0.002$ \\
\cline{2-5}
  & {w/ DropEdge} & $0.618\pm0.006$ & $0.556\pm0.006$ & $0.651\pm0.006$\\
  % & {w/ DropEdge sub} &0.562\pm0.006&0.451\pm0.015&0.470\pm0.008\\
  & {w/ DropEdge copy} & $0.593\pm0.010$ & $0.555\pm0.011$ & $0.676\pm0.005$ \\
    & {w/ DropEdge AllView} & $0.618\pm0.006$ & $0.567\pm0.011$ & $0.682\pm0.002$ \\
   & {w/ DropEdge MV} & ${\bf 0.623}\pm0.003$ & ${\bf 0.591}\pm0.006$&${\bf 0.687}\pm0.001$\\
\hline\hline%  \bottomrule
%   \bottomrule
\end{tabular}
% \end{sc}
%\end{small}
\end{center}
%\vskip -0.1in
\end{table*}

\begin{table}[ht]
\begin{center}
\caption{Mean Micro-F1 scores with standard deviations
of the mean on three real-world datasets. %datasets. 
Results are provided from %with 
runs with 10 random seeds on GAug-M.}
\label{tab:results_gaugm}
%\small 
\vspace{0.2cm}
\begin{tabular}{lcccc}
%\toprule
\multicolumn{1}{c}{\bf BACKBONE}  &\multicolumn{1}{c}{\bf METHOD} &\multicolumn{1}{c}{\bf PPI-BP}  &\multicolumn{1}{c}{\bf HPO-METAB}&\multicolumn{1}{c}{\bf HPO-NEURO}\\
\hline
%\midrule
\multirow{5}*{GLASS}  & {Original} & $0.610\pm0.006$  & $0.600\pm0.003$  & $0.678\pm0.004$ \\
  \cline{2-5}
& {w/ GAug-M} &$0.609\pm0.003$& $0.596\pm0.007$ & ${0.681}\pm0.003$\\
% & {w/ GAug-M sub} & 0.629\pm0.005&
% {\bf 0.615}\pm0.008 & 0.685\pm0.004 \\ 
& {w/ GAug-M copy} & $0.621\pm0.006$ & $0.593\pm0.016$ & $0.679\pm0.003$ \\
    & {w/ GAug-M AllView} & $0.617\pm0.004$ & $0.579\pm0.005$ & $0.672\pm0.006$ \\
& {w/ GAug-M MV} &${\bf 0.625}\pm0.004$& ${\bf 0.598}\pm0.005$ & ${\bf 0.682}\pm0.003$ \\
\hline %\bottomrule
\multirow{7}*{GCN}  & {Original} & $0.613\pm0.008$ & $0.553\pm0.018$ & $0.658\pm0.007$  \\
 \cline{2-5}
& {w/ GAug-M} &$0.603\pm0.008$& $0.557\pm0.011$&$0.641\pm0.006$\\
% & {w/ GAug-M sub} &0.609\pm0.010&
% 0.567\pm0.009&0.530\pm0.013\\
& {w/ GAug-M copy} & $0.556\pm0.007$ & $0.335\pm0.025$ & $0.306\pm0.016$ \\
& {w/ GAug-M AllView} & $0.591\pm0.006$ & $0.544\pm0.008$ & $0.593\pm0.007$ \\
& {w/ GAug-M MV} &${\bf 0.616}\pm0.008$& ${\bf 0.564}\pm0.008$ & $0.640\pm0.003$ \\
\hline %\bottomrule
\multirow{7}*{GSAGE}  & {Original} & $0.621\pm0.006$ & $0.581\pm0.008$ & $0.684\pm0.002$ \\
% \cline{2-5}
%   & {w/ DropEdge} &
%  0.618\pm0.006 & 0.556\pm0.006 & 0.651\pm0.006 \\
%   & {w/ DropEdge sub} & 0.583\pm0.006 & 0.584\pm0.009 & 0.651\pm0.005 \\
%    % & {w/ DropEdge whole} &  {\bf 0.884}\pm0.011 & {\bf 0.593}\pm0.007 & {\bf 0.692}\pm0.002 & 0.613\pm0.006\\
%    & {w/ DropEdge MV} & {\bf 0.623}\pm0.003 & {\bf 0.591}\pm0.006 & {\bf 0.687}\pm0.001 & \\
\cline{2-5}
& {w/ GAug-M} & $0.603\pm0.017$& $0.588\pm0.012$ & $0.672\pm0.004$ \\
% & {w/ GAug-M sub} &0.609\pm0.010&
% 0.567\pm0.009&0.530\pm0.013\\
& {w/ GAug-M copy} & $0.606\pm0.006$ & $0.546\pm0.007$ & $0.684\pm0.003$ \\
& {w/ GAug-M AllView} & $0.622\pm0.005$ & $0.583\pm0.001$ & $0.682\pm0.003$ \\
& {w/ GAug-M MV} &${\bf 0.628}\pm0.005$& ${\bf 0.591}\pm0.008$ &${\bf 0.689}\pm0.003$ \\
\hline\hline %\bottomrule
\end{tabular}
\end{center}
\end{table}

\textbf{On augmentation on subgraphs}
The augmentation should take place in the $k$-hop neighbors of a subgraph to avoid the requirement of extreme fine tunes on hyperparameters, as demonstrated from Tables~\ref{tab:results_dropedge} and \ref{tab:results_gaugm}. Comparing fields of
\textbf{w/ Augmentation} with \textbf{ w/ Augmentation MV} we can see that the performance of our approach at least maintains the performance of implementing augmentation on whole graph. On the GSAGE backbone, we can see that DropEdge and GAug-M promote the performance of HPO-NEURO by 3.6\% and 2.7\%, respectively, compared with the direct augmentation. This result shows that our approach is able to get stronger results within a limited hyperparameter space because we perform the augmentation on neighbor regions, which is a more important region for augmentation.

\textbf{On creating subgraph multi-view only}
As discussed in Section~\ref{subgraphormulti}, one way to perform augmentation is to duplicate a graph. From Tables~\ref{tab:results_dropedge} and \ref{tab:results_gaugm}, comparing fields of  
\textbf{w/ Augmentation copy} with \textbf{w/ Augmentation MV}, our approach always performs better than copying the original graph. We even observe a sharp decrease by over 21.8\% on the GCN backbone with HPO datasets. This result indicates the importance of the addition of the contrastive loss, and proves that duplicating the original graph may take potential failures in aligning the original graph and the augmented graph.

\textbf{On preserving one view} 
The argument that augmenting all views can bring more training support to models but lose the information from the original graph, as discussed in Section~\ref{subgraphormulti} can also be provided by Tables~\ref{tab:results_dropedge} and \ref{tab:results_gaugm}. %From Table \ref{tab:results_dropedge}, \ref{tab:results_gaugm}, 
Comparing fields of \textbf{w/ Augmentation AllView} with \textbf{ w/ Augmentation MV}, we can find that %the result of 
our approach also brings some improvements. These results echo that it's necessary to keep at least one view to make the subgraphs lossless in information.
\textbf{w/ Augmentation copy} with \textbf{ w/ Augmentation MV} field,

\subsubsection{Ablation Study on the Contrastive Loss}

{
% \color{red}{
In addition to the comparison on different augmentation strategies,}
 we also perform %did 
the ablation on the importance of the contrastive loss, as reported in Tables~\ref{tab:ContrastiveDropEdge} and \ref{tab:ContrastiveGAug}. % report the results. 
Overall, the results show that the contrastive loss can greatly improve the learning process of subgraph embeddings, which utilizes maximizing the agreement between the original subgraph and the augmented subgraph. Also, the decrease on \textbf{w/ Augmentation copy} compared with our approach echoes this point of view.

% \chengwei{Argument A (such as the advantagous of multi view) can be proved/demonstrated by XXX[experimental results]. Discuss and analysis results, indicate balabala, then talk some potential reason. Therefore, xxx is advantageous than XXX}

\begin{table}[ht]
\begin{center}
%\begin{small}
\caption{Ablation Study on DropEdge.}
\label{tab:ContrastiveDropEdge}
\vspace{0.2cm}
\begin{tabular}{lcccccccc}
%\toprule
\multicolumn{1}{c}{\bf BACKBONE}  &\multicolumn{1}{c}{\bf METHOD} &\multicolumn{1}{c}{\bf PPI-BP}  &\multicolumn{1}{c}{\bf HPO-METAB}&\multicolumn{1}{c}{\bf HPO-NEURO} & \\ 
\hline
%\midrule
\multirow{3}*{GLASS}  & {Original} &
 $0.610\pm0.006$  & $0.600\pm0.003$  & $0.678\pm0.004$  \\
  \cline{2-5}
& {w/  No Contrast} & $0.618\pm0.006$ & $0.597\pm0.005$ & ${\bf 0.685}\pm0.003$ \\
& {w/  Contrast} &${\bf 0.628}\pm0.005$ & ${\bf 0.607}\pm0.008$ & $0.683\pm0.004$ \\
\hline %\bottomrule
\multirow{3}*{GCN}  & {Original} & $0.613\pm0.008$ & $0.553\pm0.018$ & $0.658\pm0.007$  \\
  \cline{2-5}
& {w/ No Contrast} & $0.616\pm0.006$ & $0.506\pm0.030$ & $0.606\pm0.011$ \\
& {w/ Contrast} & ${\bf 0.619}\pm0.006$ & ${\bf 0.567}\pm0.006$ & ${\bf 0.622}\pm0.007$ \\
\hline %\bottomrule
\multirow{3}*{GSAGE}  & {Original} & $0.621\pm0.006$ & $0.581\pm0.008$ & $0.684\pm0.002$ \\
  \cline{2-5}
& {w/ No Contrast} &$0.616\pm0.007$&  $0.583\pm0.008$ &$0.684\pm0.004$ \\
& {w/ Contrast} & ${\bf 0.623}\pm0.003$&${\bf 0.591}\pm0.006$&${\bf 0.687}\pm0.001$\\
\hline\hline %\bottomrule
% \bottomrule
\end{tabular}
%\end{small}
\end{center}
\end{table}

\begin{table}[ht]
%\small 
\begin{center}
\caption{Ablation Study on GAug-M.}
\label{tab:ContrastiveGAug} 
\vspace{0.2cm}
\begin{tabular}{lccccc}
%\toprule
\multicolumn{1}{c}{\bf BACKBONE}  &\multicolumn{1}{c}{\bf METHOD} &\multicolumn{1}{c}{\bf PPI-BP}  &\multicolumn{1}{c}{\bf HPO-METAB}&\multicolumn{1}{c}{\bf HPO-NEURO}\\
\hline
%\midrule
\multirow{3}*{GLASS}  & {Original} & $0.610\pm0.006$ & $0.600\pm0.003$  & $0.678\pm0.004$  \\
  \cline{2-5}
& {w/  No Contrast} & $0.611\pm0.006$ & $0.591\pm0.007$ & ${\bf 0.682}\pm0.003$ \\
& {w/  Contrast} &${\bf 0.625}\pm0.004$& ${\bf 0.598}\pm0.005$ & $0.680\pm0.003$ & \\
\hline %\bottomrule
\multirow{3}*{GCN}  & {Original} & $0.613\pm0.008$ & $0.553\pm0.018$ & $0.658\pm0.007$  \\
  \cline{2-5}
& {w/  No Contrast} & $0.613\pm0.008$ & $0.513\pm0.019$ & $0.580\pm0.015$ \\
& {w/  Contrast} &${\bf 0.616}\pm0.008$& ${\bf 0.564}\pm0.008$ & ${\bf 0.640}\pm0.003$ \\
\hline %\bottomrule
\multirow{3}*{GSAGE}  & {Original} &  $0.621\pm0.006$ & $0.581\pm0.008$ & $0.684\pm0.002$ \\
  \cline{2-5}
& {w/  No Contrast} &$0.621\pm0.005$ & $0.582\pm0.008$ & $0.684\pm 0.003$ \\
& {w/  Contrast} &$ {\bf 0.628}\pm0.005$ & ${\bf 0.591}\pm 0.008$ &${\bf 0.689}\pm 0.003$ \\
\hline %\bottomrule
\end{tabular}
\end{center}
\end{table}

\subsubsection{Computational Time}
% \begin{wraptable}[6]{r}{0.62\textwidth}
% \begin{minipage}{0.60\textwidth}
% \vspace{-0.8cm}
\begin{wraptable}[7]{r}{0.45\textwidth}
\begin{minipage}{0.43\textwidth}
% \begin{table}[ht]
% \scalebox{0.7}{
%\small 
\begin{center}
\vspace{-0.8cm}

\caption{The training time records generated by the experiments. }
\label{tab:Complexity} 
\vspace{-0.2cm}
\begin{tabular}{lccccc}
\multicolumn{1}{c}{\bf METHOD}  &\multicolumn{1}{c}{\bf TIME / EPOCH (s)}   \\
\hline
w/ GAug-M & 0.439 \\
% \midrule
\textbf{w/ GAug-M MV} & \textbf{0.465}\\
% \midrule
w/ GAug-M copy & 0.788 \\
\hline

\end{tabular}

\end{center}

% \end{table}
\end{minipage}
% \vspace{}
\end{wraptable}
% \end{verbatim}
We compare the computational time for % We did a computational time comparison on  
the direct augmentation on original graphs {\bf w/ Augmentation}, copying the original graph {\bf w/ Augmentation Copy} and our approach {\bf w/ Augmentation MV}, based on the same experimental settings and hyperparameters. 
We generate the records %The records are generated by 
using GAug-M as the augmentation approach, GLASS as the backbone, and PPI-BP as the dataset. The results are shown in Table~\ref{tab:Complexity}. We can see that using multi-view does not greatly increase the computational complexity and the memory complexity in practice. Meanwhile, directly duplicating the graph will almost double the training time, which means an unaffordable increase on the computational complexity. This result shows that our approach gains both efficiency and effectiveness in subgraph augmentation comparing with other baseline approaches.
%\yili{TBD}

\section{Conclusion}

We propose a novel model-agnostic subgraph augmentation strategy to facilitate subgraph-based GNNs. By creating a new subgraph and link to the original graph, it will include more diversity in message passing to the graph to enhance model training support. This subgraph-specific  augmentation strategy can improve the performance and the robustness of a graph neural network. A bunch of improved experiments on both GAug-M and DropEdge on different datasets show the generalizability on models and augmentation strategies. It's expected that our research will empower the subgraph representation learning to go further and broader.

\bibliography{iclr2023_conference}

\begin{thebibliography}{23}
\providecommand{\natexlab}[1]{#1}
\providecommand{\url}[1]{\texttt{#1}}
\expandafter\ifx\csname urlstyle\endcsname\relax
  \providecommand{\doi}[1]{doi: #1}\else
  \providecommand{\doi}{doi: \begingroup \urlstyle{rm}\Url}\fi

\bibitem[Alsentzer et~al.(2020)Alsentzer, Finlayson, Li, and
  Zitnik]{alsentzer2020subgraph}
Emily Alsentzer, Samuel~G. Finlayson, Michelle~M. Li, and Marinka Zitnik.
\newblock Subgraph neural networks.
\newblock In \emph{Advances in Neural Information Processing Systems}, 2020.
\newblock URL \url{https://openreview.net/forum?id=s7kdBmjRn4}.

\bibitem[Fan et~al.(2020)Fan, Wang, Shi, Lu, Lin, and Wang]{o2mac}
Shaohua Fan, Xiao Wang, Chuan Shi, Emiao Lu, Ken Lin, and Bai Wang.
\newblock {One2Multi} graph autoencoder for multi-view graph clustering.
\newblock In \emph{Proceedings of The Web Conference 2020 (WWW ’20)}, 2020.

\bibitem[Gilmer et~al.(2017)Gilmer, Schoenholz, Riley, Vinyals, and
  Dahl]{DBLP:journals/corr/GilmerSRVD17}
Justin Gilmer, Samuel~S. Schoenholz, Patrick~F. Riley, Oriol Vinyals, and
  George~E. Dahl.
\newblock Neural message passing for quantum chemistry.
\newblock In \emph{International Conference on Machine Learning}, volume~70,
  pp.\  1263–1272, 2017.

\bibitem[Hamilton et~al.(2017)Hamilton, Ying, and
  Leskovec]{hamilton2017inductive}
Will Hamilton, Zhitao Ying, and Jure Leskovec.
\newblock Inductive representation learning on large graphs.
\newblock volume~30, pp.\  1024--1034, 2017.

\bibitem[Hartley et~al.(2020)Hartley, Lemire, Kernohan, Howley, Adams, and
  Boycott]{hartley2020new}
Taila Hartley, Gabrielle Lemire, Kristin~D. Kernohan, Heather~E. Howley,
  David~R. Adams, and Kym~M. Boycott.
\newblock New diagnostic approaches for undiagnosed rare genetic diseases.
\newblock \emph{Annu. Rev. Genomics. Hum. Genet.}, 21\penalty0 (1):\penalty0
  351--372, 2020.

\bibitem[Hassani \& Khasahmadi(2020)Hassani and
  Khasahmadi]{hassani2020contrastive}
Kaveh Hassani and Amir~Hosein Khasahmadi.
\newblock Contrastive multi-view representation learning on graphs.
\newblock In \emph{International Conference on Machine Learning}, volume 385,
  pp.\  4116--4126, 2020.

\bibitem[Hastie et~al.(2009)Hastie, Tibshirani, and
  Friedman]{hastie2009elements}
Trevor Hastie, Robert Tibshirani, and Jerome~H. Friedman.
\newblock \emph{The Elements of Statistical Learning: Data Mining, Inference,
  and Prediction}, volume~2.
\newblock Springer, 2009.

\bibitem[Jiao et~al.(2020)Jiao, Xiong, Zhang, Zhang, Zhang, and
  Zhu]{jiao2020sub}
Yizhu Jiao, Yun Xiong, Jiawei Zhang, Yao Zhang, Tianqi Zhang, and Yangyong Zhu.
\newblock Sub-graph contrast for scalable self-supervised graph representation
  learning.
\newblock In \emph{2020 IEEE International Conference on Data Mining}, pp.\
  222--231, 2020.

\bibitem[Kim \& Oh(2022)Kim and Oh]{kim2022efficient}
Dongkwan Kim and Alice Oh.
\newblock Efficient representation learning of subgraphs by subgraph-to-node
  translation.
\newblock In \emph{International Conference on Learning Representations 2022
  Workshop on Geometrical and Topological Representation Learning}, 2022.
\newblock URL \url{https://openreview.net/forum?id=BgLaE-k6gc}.

\bibitem[Kipf \& Welling(2016)Kipf and Welling]{kipf2016variational}
Thomas~N Kipf and Max Welling.
\newblock Variational graph auto-encoders.
\newblock \emph{arXiv preprint arXiv:1611.07308}, 2016.

\bibitem[Kipf \& Welling(2017)Kipf and Welling]{kipf2016semi}
Thomas~N. Kipf and Max Welling.
\newblock Semi-supervised classification with graph convolutional networks.
\newblock In \emph{International Conference on Learning Representations}, 2017.
\newblock URL \url{https://openreview.net/forum?id=SJU4ayYgl}.

\bibitem[Ma et~al.(2020)Ma, Bian, Rong, Huang, Xu, Xie, Ye, and
  Huang]{DBLP:journals/corr/abs-2005-13607}
Hehuan Ma, Yatao Bian, Yu~Rong, Wenbing Huang, Tingyang Xu, Weiyang Xie, Geyan
  Ye, and Junzhou Huang.
\newblock Multi-view graph neural networks for molecular property prediction.
\newblock \emph{arXiv preprint arXiv:2005.13607}, 2020.

\bibitem[Meng et~al.(2018)Meng, Mouli, Ribeiro, and Neville]{meng2018subgraph}
Changping Meng, S~Chandra Mouli, Bruno Ribeiro, and Jennifer Neville.
\newblock Subgraph pattern neural networks for high-order graph evolution
  prediction.
\newblock In \emph{Proceedings of the AAAI Conference on Artificial
  Intelligence}, volume~32, 2018.

\bibitem[Rong et~al.(2020)Rong, Huang, Xu, and Huang]{rong2020dropedge}
Yu~Rong, Wenbing Huang, Tingyang Xu, and Junzhou Huang.
\newblock Dropedge: Towards deep graph convolutional networks on node
  classification.
\newblock In \emph{International Conference on Learning Representations}, 2020.
\newblock URL \url{https://openreview.net/forum?id=Hkx1qkrKPr}.

\bibitem[Splinter et~al.(2018)Splinter, Adams, Bacino, Bellen, Bernstein,
  Cheatle-Jarvela, Eng, Esteves, Gahl, Hamid, Jacob, Kikani, Koeller, Kohane,
  Lee, Loscalzo, Luo, McCray, Metz, Mulvihill, Nelson, Palmer, Phillips, Pick,
  Postlethwait, Reuter, Shashi, Sweetser, Tifft, Walley, Wangler, Westerfield,
  Wheeler, Wise, Worthey, Yamamoto, and Ashley]{splinter2018effect}
Kimberly Splinter, David~R. Adams, Carlos~A. Bacino, Hugo~J. Bellen,
  Jonathan~A. Bernstein, Alys~M. Cheatle-Jarvela, Christine~M. Eng, Cecilia
  Esteves, William~A. Gahl, Rizwan Hamid, Howard~J. Jacob, Bijal Kikani,
  David~M. Koeller, Isaac~S. Kohane, Brendan~H. Lee, Joseph Loscalzo, Xi~Luo,
  Alexa~T. McCray, Thomas~O. Metz, John~J. Mulvihill, Stanley~F. Nelson,
  Christina~G.S. Palmer, John~A. Phillips, Leslie Pick, John~H. Postlethwait,
  Chloe Reuter, Vandana Shashi, David~A. Sweetser, Cynthia~J. Tifft, Nicole~M.
  Walley, Michael~F. Wangler, Monte Westerfield, Matthew~T. Wheeler,
  Anastasia~L. Wise, Elizabeth~A. Worthey, Shinya Yamamoto, and Euan~A. Ashley.
\newblock Effect of genetic diagnosis on patients with previously undiagnosed
  disease.
\newblock \emph{N. Engl. J. Med.}, 379\penalty0 (22):\penalty0 2131--2139,
  2018.

\bibitem[Veli\v{c}kovi\'{c} et~al.(2019)Veli\v{c}kovi\'{c}, Fedus, Hamilton,
  Li\`{o}, Bengio, and Hjelm]{velivckovic2018deep}
Petar Veli\v{c}kovi\'{c}, William Fedus, William~L. Hamilton, Pietro Li\`{o},
  Yoshua Bengio, and R.~Devon Hjelm.
\newblock Deep graph infomax.
\newblock In \emph{International Conference on Learning Representations}, 2019.
\newblock URL \url{https://openreview.net/forum?id=rklz9iAcKQ}.

\bibitem[Wang \& Zhang(2021)Wang and Zhang]{wang2021glass}
Xiyuan Wang and Muhan Zhang.
\newblock {GLASS}: {GNN} with labeling tricks for subgraph representation
  learning.
\newblock In \emph{International Conference on Learning Representations}, 2021.
\newblock URL \url{https://openreview.net/forum?id=XLxhEjKNbXj}.

\bibitem[You et~al.(2020)You, Chen, Sui, Chen, Wang, and Shen]{you2020graph}
Yuning You, Tianlong Chen, Yongduo Sui, Ting Chen, Zhangyang Wang, and Yang
  Shen.
\newblock Graph contrastive learning with augmentations.
\newblock In H.~Larochelle, M.~Ranzato, R.~Hadsell, M.F. Balcan, and H.~Lin
  (eds.), \emph{Advances in Neural Information Processing Systems}, volume~33,
  pp.\  5812--5823, 2020.

\bibitem[Zhang et~al.(2021)Zhang, Li, Xia, Wang, and Jin]{zhang2021labeling}
Muhan Zhang, Pan Li, Yinglong Xia, Kai Wang, and Long Jin.
\newblock Labeling trick: A theory of using graph neural networks for
  multi-node representation learning.
\newblock In A.~Beygelzimer, Y.~Dauphin, P.~Liang, and J.~Wortman Vaughan
  (eds.), \emph{Advances in Neural Information Processing Systems}, 2021.
\newblock URL \url{https://openreview.net/forum?id=Hcr9mgBG6ds}.

\bibitem[Zhao et~al.(2021)Zhao, Liu, Neves, Woodford, Jiang, and
  Shah]{zhao2021data}
Tong Zhao, Yozen Liu, Leonardo Neves, Oliver Woodford, Meng Jiang, and Neil
  Shah.
\newblock Data augmentation for graph neural networks.
\newblock In \emph{Proceedings of the AAAI Conference on Artificial
  Intelligence}, volume~35, pp.\  11015--11023, 2021.

\bibitem[Zheng et~al.(2020)Zheng, Zong, Cheng, Song, Ni, Yu, Chen, and
  Wang]{zheng2020robust}
Cheng Zheng, Bo~Zong, Wei Cheng, Dongjin Song, Jingchao Ni, Wenchao Yu, Haifeng
  Chen, and Wei Wang.
\newblock Robust graph representation learning via neural sparsification.
\newblock In \emph{International Conference on Machine Learning}, volume 119,
  pp.\  11458--11468, 2020.

\bibitem[Zhu et~al.(2020)Zhu, Xu, Yu, Liu, Wu, and Wang]{zhu2020deep}
Yanqiao Zhu, Yichen Xu, Feng Yu, Qiang Liu, Shu Wu, and Liang Wang.
\newblock Deep graph contrastive representation learning.
\newblock \emph{arXiv preprint arXiv:2006.04131}, 2020.

\bibitem[Zitnik et~al.(2018)Zitnik, Sosi\v{c}, Maheshwari, and
  Leskovec]{biosnapnets}
Marinka Zitnik, Rok Sosi\v{c}, Sagar Maheshwari, and Jure Leskovec.
\newblock {BioSNAP Datasets}: {Stanford} biomedical network dataset collection,
  2018.
\newblock URL \url{http://snap.stanford.edu/biodata}.

\end{thebibliography}
\bibliographystyle{iclr2023_conference}

\appendix
\section{Appendix}
\label{sec:appendix}
% \zhou{You may include other additional sections here.}
\subsection{Implementation details}
\subsubsection{Hyperparameter tuning}
By solidating the hyperparameters by GLASS on all backbone settings, we implement the grid hyperparameter searching on augmentations. 
For  $\lambda$, we select a searching space \{0, 0.01, 0.05, 0.1, 0.5, 1,1.25, 1.5, 2, 2.5\}.
For DropEdge-related tasks, our searching space for $p$ is \{0.1, 0.2, 0.3, 0.4, 0.5\}.
For GAug-related tasks, our searching space for (rm\_pct, add\_pct) is \{(0, 50), (15, 35), (30, 20), (45, 5)\}. Since we used VGAE to generate different probabilistic edge distribution, we selected $i$ from \{1, 2\}.
\end{document}